\documentclass[acmlarge]{acmart}

\AtBeginDocument{%
  }

\usepackage{booktabs}%
\usepackage{graphicx}%
\usepackage{multirow}%
\usepackage{amsmath,amsfonts}%
\usepackage{mathrsfs}%
\usepackage[title]{appendix}%
\usepackage{xcolor}%
\usepackage{textcomp}%
\usepackage{manyfoot}%
\usepackage{algorithm}%
\usepackage{algorithmicx}%
\usepackage{algpseudocode}%
\usepackage{listings}%
\usepackage{comment}
\usepackage{optidef}
\usepackage{subfigure}
\usepackage{color}
\usepackage{url}
\usepackage{hyperref}

\begin{document}

\title{A Self-Adaptive Penalty Method for Integrating Prior Knowledge Constraints into Neural ODEs}

\author{C. Coelho}
\email{cmartins@cmat.uminho.pt}
\orcid{0009-0009-4502-937X}
\author{M. Fernada P. Costa}
\orcid{0000-0001-6235-286X}
\email{mfc@math.uminho.pt}
\affiliation{%
  \institution{Centre of Mathematics (CMAT), University of Minho}
  \city{Braga}
  \country{Portugal}
  \postcode{4710-57}
}

\author{L.L. Ferrás}
\email{lferras@fe.up.pt}
\orcid{0000-0001-5477-3226}
\affiliation{
    \institution{Centre of Mathematics (CMAT), University of Minho}
  \city{Braga}
  \country{Portugal}
  \postcode{4710-057}
  }
\affiliation{%
  \institution{Department of Mechanical Engineering (Section of Mathematics) - FEUP, University of Porto}
  \city{Porto}
  \country{Portugal}
  \postcode{4200-465}
  }

\renewcommand{\shortauthors}{C. Coelho et al.}
\acmArticleType{Research}
\acmCodeLink{https://github.com/borisveytsman/acmart}
\acmDataLink{htps://zenodo.org/link}
\acmContributions{All authors contributed equally to the manuscript.}

\begin{abstract}
The continuous dynamics of natural systems has been effectively modelled using Neural Ordinary Differential Equations (Neural ODEs). However, for accurate and meaningful predictions, it is crucial that the models follow the underlying rules or laws that govern these systems.
In this work, we propose a self-adaptive penalty algorithm for Neural ODEs to enable modelling of constrained natural systems. The proposed self-adaptive penalty function can dynamically adjust the penalty parameters. The explicit introduction of prior knowledge helps to increase the interpretability of Neural ODE models.
We validate the proposed approach by modelling three natural systems with prior knowledge constraints: population growth, chemical reaction evolution, and damped harmonic oscillator motion.
The numerical experiments and a comparison with other penalty Neural ODE approaches and \emph{vanilla} Neural ODE, demonstrate the effectiveness of the proposed self-adaptive penalty algorithm for Neural ODEs in modelling constrained natural systems. Moreover, the self-adaptive penalty approach provides more accurate and robust models with reliable and meaningful predictions. The proposed method can be generalised to other neural network architectures.
The code to replicate the experiments presented in this work is available at \url{https://github.com/CeciliaCoelho/PriorKnowledgeNeuralODE} \footnote{available after acceptance}.
\end{abstract}

\maketitle

\section{Introduction}

Accurately predicting the behaviour of natural systems requires mathematical equations that describe the relationships between their variables over time. Due to its continuous-time dynamics, these systems are often formulated by Ordinary Differential Equations (ODEs). ODEs provide a robust framework for understanding and analysing the continuous changes that occur in natural systems. 

However, traditional methods of modelling natural systems using ODEs can be challenging due to the complexity of the interactions between variables. In recent years, Neural Networks (NNs) have been used to model the dynamics of natural systems based on experimental data. However, the resulting models are typically discrete and may not be suitable for modelling continuous-time dynamics. Specifically, the discrete nature of NN models limits their ability to capture the complex interactions between variables in natural systems, which are often continuous \cite{chenNeuralOrdinaryDifferential2019}. 

Neural ODEs \cite{chenNeuralOrdinaryDifferential2019} are a NN architecture that introduces the concept of adjusting an unknown continuous-time ODE system, $\boldsymbol{f_\theta}$, such that its solution curve fits the data. The function $\boldsymbol{f_\theta}$ is defined by a NN with parameters $\boldsymbol{\theta}$, and the result of training a Neural ODE is an ODE system. To make predictions, an Initial Value Problem (IVP) is solved \cite{chenNeuralOrdinaryDifferential2019}. Some natural systems modelled with Neural ODEs are reported in the literature \cite{suKineticsParameterOptimization2022,xingContinuousGlucoseMonitoring2022}.

In \cite{suKineticsParameterOptimization2022} the authors use a Neural ODE to optimise the kinetic parameters of chemical reactions, demonstrating the effectiveness of this approach in modelling complex combustion phenomena.

In \cite{xingContinuousGlucoseMonitoring2022} the authors propose using a Neural ODE to model glucose levels based on sparse training data, which exhibits high precision in predicting blood glucose trends, demonstrating its potential for practical applications in patient care.


Natural systems often have governing laws that are mathematically expressed as constraints. However, when we create mathematical models for these systems, we may encounter complex and nonlinear dependencies among variables, which can make the modelling process difficult and lead to inaccurate predictions.


NNs are commonly referred to as "black-box" models because they are complex and opaque. It can be challenging to understand how the model uses the training data to make predictions, which raises concerns about their interpretability and reliability. This lack of transparency can lead to distrust in the scientific community, as there is no guarantee that NN-based models satisfy the underlying governing laws of these natural systems.

Incorporating prior knowledge into NNs can reduce the need for large amounts of training data, improve generalisation, predictive performance and reduce the risk of overfitting.

In the literature, several strategies have been proposed to incorporate prior knowledge constraints into NNs \cite{vonruedenInformedMachineLearning2023}. A popular and well-established approach for constrained optimisation is the use of penalty methods, that convert the constrained problem into an unconstrained one via a penalty function. The penalty function combines the objective function and penalty terms that penalise the constraints violations. Penalty methods have already been used in NNs and Neural ODEs \cite{daiNovelEstimationMethod2019,tuorConstrainedNeuralOrdinary2020}.

For instance in \cite{daiNovelEstimationMethod2019} the state of health of a lithium-ion battery is predicted by introducing constraints related to the charging profile of the battery. 
In \cite{tuorConstrainedNeuralOrdinary2020} the authors introduce prior knowledge to guarantee safety and performance when modelling industrial systems.
In both papers a quadratic penalty function is used.

We note that, penalty methods are very straightforward to implement and easy to incorporate into standard NN training algorithms. However, a major challenge is the choice of the penalty parameter $\mu$. The penalty parameter plays a crucial role in controlling the trade-off between the original loss function and the penalty term, influencing the overall training process and the model's final performance. Thus, selecting an appropriate value for $\mu$ is not a trivial task, as it requires a careful balance.
To address the issue of selecting an appropriate $\mu$, adaptive penalty functions have been proposed in the field of constrained optimisation. These functions are designed to dynamically adjust penalty parameters during the optimisation process \cite{aliPenaltyFunctionbasedDifferential2013,costaTheoreticalPracticalConvergence2017}. 

In \cite{aliPenaltyFunctionbasedDifferential2013} the authors propose an adaptive penalty function-based differential evolution algorithm to solve constrained global optimisation problems, which relies on a parameter-free adaptive penalty function. In \cite{costaTheoreticalPracticalConvergence2017} the authors develop a self-adaptive penalty function. This penalty function is designed to dynamically adjust the parameters of the dynamic process. The authors integrate it with a population-based meta-heuristic, called hybrid self-adaptive penalty firefly algorithm. This hybrid algorithm is designed specifically to address challenges posed by nonsmooth nonconvex constrained global optimisation problems.


This work presents a new approach to modelling constrained natural systems using Neural ODEs. We introduce a new self-adaptive penalty function and algorithm that adjusts the penalty parameters $\mu$ based on constraints violation (at each iteration).  Furthermore, the proposed methods are specifically designed to be used effectively in training NNs, enhancing their performance in constrained scenarios. Our approach is an improvement over traditional penalty methods which require an appropriate initialisation and tuning of penalty parameters $\mu$ during the optimisation process. This selection is challenging and time-consuming.

The proposed method applies stronger penalties to heavily violated constraints and weaker penalties to slightly violated ones, balancing the objective function minimisation with constraint satisfaction. By adapting $\mu$ over time, our approach is more robust and adaptable to possible changes in input data distribution. Our contributions are significant steps towards more effective and flexible modelling of constrained natural systems using Neural ODEs.

To the best of our knowledge this is the first time a self-adaptive penalty function is proposed to incorporate prior knowledge constraints into a NN architecture, more specifically into a Neural ODE.

This paper is organised as follows. Section \ref{sec:background} provides a brief introduction to Neural ODEs and to the formulation of constrained optimisation problems. Section \ref{sec:method} presents the proposed constrained time series optimisation problems formulation followed by the new self-adaptive penalty function and finishing with the self-adaptive penalty algorithm for Neural ODEs. Section \ref{sec:experiments} presents the results obtained for some numerical experiments. The conclusions are presented in Section \ref{sec:conclusion}. Additionally, a detailed description of a dataset developed and introduced for the first time in this work is included in Appendix \ref{app:DHO}.

\section{Background} \label{sec:background}

In this section we give a brief introduction to the concept of Neural ODEs and we present the redefinition of a constrained optimisation problem applied to time series data.

\subsection{Neural ODEs}

Consider a time series describing a natural system. Let $\boldsymbol{Y}=(\boldsymbol{y}_0, \boldsymbol{y}_2, \dots, \boldsymbol{y}_{N-1})$ be the corresponding ground-truth time series, with $\boldsymbol{y}_n \in \mathbb{R}^{d^*}$ and $\boldsymbol{\hat{Y}}=(\boldsymbol{\hat{y}}_0, \boldsymbol{\hat{y}}_2, \dots, \boldsymbol{\hat{y}}_{N-1})$ the prediction, with $\boldsymbol{\hat{y}}_{n} \in \mathbb{R}^{d^*}$, at time step $t_n $.

Neural ODEs are a NN architecture that adjusts a continuous-time function (ODE) to the dynamics of the training data \cite{chenNeuralOrdinaryDifferential2019}. 
A Neural ODE is composed of two components, a NN that builds the ODE dynamics $\boldsymbol{f}_{\boldsymbol{\theta}}$, of the system under study, and a numerical ODE solver. During training, the parameters $\boldsymbol{\theta}$ of the NN are adjusted by comparing the ground-truth values with the predictions made by using the ODE solver to solve an IVP over a time interval, $(t_0, t_{N-1})$:

\begin{equation*}
   \{\boldsymbol{\hat{y}}_{n}(\boldsymbol{\theta})\}_{n=0 \dots {N-1}} = ODESolve(\boldsymbol{f_\theta}, \boldsymbol{y}_0 , (t_0, t_{N-1})), 
\end{equation*}

\noindent where $(\boldsymbol{y}_0, t_0)$ is the initial condition of the IVP, $\boldsymbol{y}(t_0)=\boldsymbol{y}_0$, \, $\boldsymbol{\hat{y}}_{n}(\boldsymbol{\theta})$ is the solution at an arbitrary time step $t_n$ with $n=0, \dots, N-1$. 
Therefore, the solver produces predictions that are adjusted by the NN to be as close as possible to the ground-truth data $\boldsymbol{Y}$.  

%
%
%
%
%

\subsection{Constrained Optimisation Problem}

A time series can be given by a matrix wherein each time step corresponds to a vector of state variables' values that encapsulate the system's evolution over time. When dealing with a constrained natural system, it becomes imperative that these constraints are consistently satisfied at every time step. Thus, the problem of estimating the parameters $\boldsymbol{\theta}$ of the NN, that builds the ODE dynamics of a constrained natural system, must be formulated in align with this temporal constraint adherence.
Therefore, a formulation for this type of problem is as follows:

\begin{mini}|l|[0]
    {\boldsymbol{\theta} \in \mathbb{R}^{n_\theta}}{l(\boldsymbol{\theta}) = \frac{1}{N}\sum_{n=0}^{N-1} ( \boldsymbol{\hat{y}}_{n}(\boldsymbol{\theta})-\boldsymbol{y}_{n} )^2}
    {\label{eq:constrainedMinProblem}}
    {}
    \addConstraint{c_{t_n}^i(\boldsymbol{\hat{y}}_n(\boldsymbol{\theta}))}{= 0}, \,\,\, i \in \varepsilon  ,\,\,\, n=0,\dots,N-1
    \addConstraint{c_{t_n}^j(\boldsymbol{\hat{y}}_n(\boldsymbol{\theta}))}{ \le 0}, \,\, j \in \mathcal{I} ,\,\,\, n=0,\dots,N-1,
\end{mini}

\noindent where $l: \mathbb{R}^{n_\theta} \rightarrow \mathbb{R}$ is the loss function, $c_{t_n}^i, c_{t_n}^j: \mathbb{R}^{n_\theta} \rightarrow \mathbb{R}$ are the equality and inequality constraint functions, respectively, with $\varepsilon$ the equality and $\mathcal{I}$ inequality index sets of constraints, over the time interval $(t_0, t_{N-1})$. The set of points that satisfy all the constraints defines the feasible set $$\mathcal{S}=\{ \boldsymbol{\theta}  \in \mathbb{R}^{n_\theta} : c_{t_n}^i(\boldsymbol{\hat{y}}_n(\boldsymbol{\theta}))=0, \,  i \in \varepsilon ;\,\, c_{t_n}^j(\boldsymbol{\hat{y}}_n(\boldsymbol{\theta})) \leq 0, \, j \in \mathcal{I}, n=0,\dots,N-1 \}.$$

\textbf{Remark:} $l(\boldsymbol{\theta})$ can be any arbitrary function, that measures the error between the ground-truth and predicted values. In this work,  $l(\boldsymbol{\theta})$ is defined by the Mean Squared Error (MSE) function.

 The constrained optimisation problem \eqref{eq:constrainedMinProblem} can be rewritten and solved as an unconstrained problem using a penalty method. Penalty methods combine the objective function and the violation of each constraint  into a penalty function \cite{fletcherBriefHistoryFilter}. 

In this work, we extend penalty methods for solving the constrained problem \eqref{eq:constrainedMinProblem} which is defined by time series data.  The distinctive feature is the need for computing each constraint violation for each time step $t_n$ within the time series, yielding a vector of values. 
Subsequently, we calculate the average of these violations across all time steps to obtain a constraint violation value per constraint. For instance, the $L1$ exact penalty function for \eqref{eq:constrainedMinProblem} is defined as follows:
 
\begin{equation} \label{eq:penalty}
     L1(\boldsymbol{\theta}; \mu)=l(\boldsymbol{\theta}) + \mu P(\boldsymbol{\theta})
\end{equation}

\noindent with 

$$P(\boldsymbol{\theta})=\sum_{i \in \varepsilon} \dfrac{1}{N} \sum_{n=0}^{N-1} |c_{t_n}^i(\boldsymbol{\hat{y}}_n(\boldsymbol{\theta}))| + \sum_{j \in \mathcal{I}} \dfrac{1}{N} \sum_{n=0}^{N-1} \left([c_{t_n}^j(\boldsymbol{\hat{y}}_n(\boldsymbol{\theta}))]^+\right)$$

\noindent where $\mu > 0$ is the penalty parameter and $[z]^+$ denotes $ \max(z, 0)$. \\

Generally, the penalty function is minimised for a sequence of increasingly larger values of $\mu$ until a solution is found \cite{nocedalNumericalOptimization2006}.
Selecting an appropriate initial value of $\mu$ can be challenging. Large values of $\mu$ enforce constraints more strictly, but can result in slower convergence. Conversely, smaller values of $\mu$ penalise constraint violations more lightly, potentially leading to infeasible solutions \cite{aliPenaltyFunctionbasedDifferential2013}.
Thus, finding the right balance for $\mu$ is crucial to achieve both constraint satisfaction and efficient training. However, this process often involves a tedious trial and error approach due to the trade-off between constraint adherence and convergence speed. 

When applied to NNs, penalty terms are added into the loss function, designated hereafter by penalty loss function. The penalty loss function is defined by the original loss, $l$, and a penalty term that measures the violation of the constrained problem multiplied by a penalty parameter, $\mu P$. In practice, it is common to use a fixed value of $\mu$ during all the NN training process \cite{caiPhysicsinformedNeuralNetworks2021,daiNovelEstimationMethod2019}. However, as mentioned, selecting an appropriate value of $\mu$ is challenging and using a fixed value that remains unchanged during all the optimisation process can lead to non-feasible solutions. We note that, a fixed value of $\mu$ may not be optimal for all stages of training.  In the early stages, when the NN is far from the optimal solution, using a large $\mu$ could lead to over-penalisation of constraints, hindering the model's ability to explore the solution space effectively. On the other hand, in later stages when the NN is close to convergence, a fixed $\mu$ might not provide enough constraint enforcement, leading to sub-optimal solutions.

\section{Proposed Method}

In this section, we present the novel self-adaptive penalty function, culminating in the presentation of the self-adaptive penalty algorithm tailored specifically for Neural ODEs. 

\subsection{Self-Adaptive Penalty Function} \label{sec:method}

The aim of this study is to develop a self-adaptive penalty function, $\phi(\boldsymbol{\theta})$,
which dynamically adjusts the penalty parameters $\mu$ taking into account the constraints violation during the training process of the NN. The proposed self-adaptive penalty function adapts the penalty parameters using the information gathered from the degree of constraint violation at the current point. Thus, the proposed self-adaptive penalty function avoids the need for the user to provide values of $\mu$.

Problem \eqref{eq:constrainedMinProblem} is then rewritten as an unconstrained problem,

\begin{mini}|l|[0]
    {\boldsymbol{\theta}\in \mathbb{R}^{n_\theta}}{\phi (\boldsymbol{\theta})}
    {\label{eq:adaptive}}
    {}.
\end{mini}

To define $\phi$, the original loss function $l$ and the constraint violation values at each point $\boldsymbol{\theta}$ are normalised using a bounded function $\psi$, to ensure that all values have the same order of magnitude. In this work, we propose the following normalisation function 

\begin{equation}
\psi(x) = 1 - \dfrac{1}{1+x}
\label{eq:normalise}
\end{equation}

\noindent where $0\leq \psi(x) \leq 1$, for all $x \in \mathbb{R}_0^+$. This function was carefully designed to ensure a range of values that fosters a stable training process. During the design of the self-adaptive penalty function, other various well-known bounded functions were previously tested, such as sigmoid, hyperbolic tangent, and softmax. Nevertheless, these functions demonstrate a tendency to saturate when applied to extremely large or small input values, causing their derivatives at these points to approach zero. Consequently, this saturation behaviour can block effective learning from such inputs and potentially lead to slower or stalled learning. In contrast, \eqref{eq:normalise} exhibits dissimilar saturation tendencies for both large and small input values. This distinctive characteristic allows overcoming the issue of the derivatives approaching zero, enabling the network to continue learning from these inputs. Thus, the $\psi$ function yields enhanced performance, particularly when dealing with substantial or widely varying input values.

The loss function $l$ at each point $\boldsymbol{\theta}$ is normalised using $\psi$ given arise to the new loss function $F_{\boldsymbol{\theta}}$:

\begin{equation}
\label{eq:F}
F_{\boldsymbol{\theta}} = \psi(l(\boldsymbol{\theta})).
\end{equation}


For each equality constraint $i \in \varepsilon$, the constraint violation vector $\boldsymbol{v}^i \in \mathbb{R}^\varepsilon$, for $\boldsymbol{\theta}$, is computed for the entire time interval $(t_0, t_{N-1})$ and given by, 

$$v^i_{t_n} = |c_{t_n}^i(\boldsymbol{\hat{y}}_n(\boldsymbol{\theta})|, \,\,\, n=0,\dots,N-1.$$

Then, the normalisation $\psi$ is applied and the total violation $P_i$ is defined as the average of the violations for the entire time interval $(t_0,t_{N-1})$:

\begin{equation}
\label{eq:Pi}
P_i =  \dfrac{1}{N}  \sum_{n=0}^{N-1} \psi(v^i_{t_n}).
\end{equation}

Likewise, for each inequality constraint $j \in \mathcal{I}$ the constraint violation vector $\boldsymbol{v}^j \in \mathbb{R}^{\mathcal{I}}$, for $\boldsymbol{\theta}$, is computed for the entire time interval $(t_0, t_{N-1})$ and defined by,

$$v_{t_n}^j = [c_{t_n}^j(\boldsymbol{\hat{y}}_n(\boldsymbol{\theta}))]^+, \,\,\, n=0,\dots,N-1.$$

Then, the normalisation $\psi$ is applied and the total violation $P_j$ is given by the average of the violations for the entire time interval $(t_0,t_{N-1})$:

\begin{equation}
\label{eq:Pj}
P_j = \dfrac{1}{N} \sum_{n=0}^{N-1} \psi(v^j_{t_n}).
\end{equation}

For each equality and inequality constraints, penalty parameters $\mu_i$ and $\mu_j$, respectively, are defined by the proportion of predictions that violate the constraint in all time interval $(t_0,t_{N-1})$, at the current point:

\begin{equation}
\label{eq:mu}
\mu_i = \dfrac{\#\{t_n: v_{t_n}^i \neq 0\} }{N}, \,\,\, \mu_j = \dfrac{\#\{t_n: v_{t_n}^j \neq 0\}}{N},
\end{equation}

\noindent where $\#\{z\}$ denotes the cardinality of set $z$.

One of the advantages of computing self-adaptive penalty parameters is that it avoids the requirement of the parameters to be provided by the user. Instead, they are computed using the information, gathered at each iteration, on how many predictions violate a constraint. A constraint that is violated in a higher number of time steps (than any other constraint) will have a larger penalty parameter. Additionally, the normalisation step $\psi$ prevents numerical instability and improves the accuracy of the training process. 

Finally, the self-adaptive penalty function $\phi$ is dynamically defined at each iteration as follows. The $\phi$ function is given by the $F_{\boldsymbol{\theta}}$ function, if the current point $\boldsymbol{\theta}$ is a feasible point. Otherwise, $\phi$ is defined by $F_{\boldsymbol{\theta}}$ plus the penalty terms $P_i$ and $P_j$ multiplied by the self-adaptive penalty parameters $\mu_i$ and $\mu_j$, respectively. 

\begin{equation}
\label{eq:phi}
    \phi(\boldsymbol{\theta}) =
    \begin{cases}
        F_{\boldsymbol{\theta}}, & \text{if } \boldsymbol{\theta} \in \mathcal{S}, \cr
        F_{\boldsymbol{\theta}} + \dfrac{1}{\#\{\varepsilon\}} \sum_{i \in \varepsilon} \mu_i P_i + \dfrac{1}{\#\{\mathcal{I}\}} \sum_{j \in \mathcal{I}} \mu_j P_j , & \text{if } \boldsymbol{\theta} \notin \mathcal{S},
    \end{cases}
\end{equation}

\noindent with $F_\theta$, $P_i$, $P_j$, $\mu_i$ and $\mu_j$ given by \eqref{eq:F}-\eqref{eq:mu}, respectively.

\paragraph{Enhancement of Interpretability}

 Integrating prior knowledge constraints, such as domain-specific rules and governing laws, into NNs is a promising avenue for improving the interpretability of these models. 

 Our proposed self-adaptive penalty method presents a novel solution by eliminating the need for the user to provide values of $\mu$. Instead, our method dynamically selects and updates these parameters at each iteration of the training process. This technique not only eliminates the challenges associated with determining suitable $\mu$ values but also guarantees a balance between satisfying the imposed constraints and minimising the underlying optimisation objective. As a result, our method offers a robust assurance that the constraints will be met throughout the course of model training. Thus, the proposed self-adaptive penalty method promotes:

\begin{itemize}
    \item \textbf{Integration of Domain-Specific Knowledge:} By encoding established domain-specific rules and governing laws as constraints within the NN training method, the model is not trained solely on data but also guided by fundamental principles. This acts as a cheat to learning. In essence, this strategy can be likened to a learning shortcut, where prior knowledge acts as a compass for the model's learning process. This approach grants the model an advantage by navigating the complexities of the problem space with the aid of prior knowledge that would otherwise have to be inferred from data.
    \item \textbf{Trustworthiness:} The incorporation of prior knowledge contributes to the model's accountability. By adhering to known constraints, the model's predictions remain faithful to the governing laws of the system, mitigating the risk of generating outcomes that contradict established truths. This intrinsic alignment with domain-specific knowledge engenders trust, as the model demonstrates fidelity to the recognised rules that govern the subject matter.
    \item \textbf{Interpretable Predictions:} By incorporating prior knowledge into the models, its predictions will be made based on not only the information learnt extracted from data, that is unknown due to the "black-box" approach, but also from known valid rules/laws from the targeted system.
\end{itemize}

Thus, incorporating prior knowledge constraints into NNs bridges the gap between machine learning and the foundational laws/rules present in various domains. 
As such, the proposed method is as a valuable tool for enhancing the transparency, reliability, and interpretability of the models.

\subsection{Self-adaptive Penalty Algorithm for Neural ODEs} \label{sec:algo}


In this section, we detail the algorithm that implements the self-adaptive penalty method for Neural ODEs to model natural systems with prior knowledge constraints. The algorithm is straightforward and requires minimal modification to the traditional Neural ODE training algorithm presented in \cite{chenNeuralOrdinaryDifferential2019}.


Since in NN training it is usually employed mini-batches of data, chosen randomly, at each iteration we herein propose a \emph{best point} strategy that can be incorporated in the self-adaptive penalty algorithm. This strategy allows to control the training process by using the best point $\boldsymbol{\theta}$ generated.   

The self-adaptive penalty algorithm incorporates a \emph{best point} strategy, serving as a guide during the optimisation process. In this strategy, we keep track of the point $\boldsymbol{\theta}$ that produces the best performance and store it as $\boldsymbol{\theta}_{\text{best}}.$
If a new point does not improve upon $\boldsymbol{\theta}_{\text{best}}$, it is discarded and the optimisation process continues using $\boldsymbol{\theta}_{\text{best}}$.

\begin{definition}
   The rules to define the \emph{best point} are the following:
\begin{itemize}
    \item if both points $\boldsymbol{\theta},\boldsymbol{\theta}_{\text{best}}$ are feasible, and if $F_{\boldsymbol{\theta}}$ is lower than $F_{\text{best}}$, $\boldsymbol{\theta}$ is accepted and $\boldsymbol{\theta}_{\text{best}}$ is updated;
    \item if both points $\boldsymbol{\theta},\boldsymbol{\theta}_{\text{best}}$ are infeasible, and the new point $\boldsymbol{\theta}$ produces a constraints violation value, $P_{\boldsymbol{\theta}}$ lower than the best point $P_{\text{best}}$, $\boldsymbol{\theta}$ is accepted and $\boldsymbol{\theta}_{\text{best}}$ is updated;
    \item if both points $\boldsymbol{\theta},\boldsymbol{\theta}_{\text{best}}$ are infeasible with $P_{\boldsymbol{\theta}}=P_{\boldsymbol{\text{best}}}$, $P_{\boldsymbol{\theta}}$, and if $F_{\boldsymbol{\theta}}$ is smaller than $F_{\text{best}}$, $\boldsymbol{\theta}$ is accepted and $\boldsymbol{\theta}_{\text{best}}$ is updated;
    \item if none of the above conditions are met, the current best point is retained, and the optimisation process continues with it.
\end{itemize} 
\end{definition}

The stopping criterion is the maximum number of iterations, $k_{\max}$. When $k_{\max}$ is reached, the final parameters of the NN that build the ODE dynamics are $\boldsymbol{\theta}_{\text{best}}$.

The proposed self-adaptive penalty algorithm for Neural ODEs is presented in Algorithm \ref{alg:selfAdaptive}.

\begin{algorithm}[ht]
\caption{: The self-adaptive penalty algorithm for Neural ODEs.}
\label{alg:selfAdaptive}
\begin{algorithmic}
\State \textbf{Input:} Initial condition $(\boldsymbol{y}_0,t_0)$, start time $t_0$, end time $t_{N}$, maximum number of iterations $k_{max}$;
\State $tol \leftarrow 1e-4$;
\State $\psi(x) = 1 - \dfrac{1}{1+x}$; 
\State $\boldsymbol{f_\theta} = DynamicsNN()$;
\State Initialise parameters $\boldsymbol{\theta}$;
\State $\phi_{\text{best}} \leftarrow +\infty$; 
\State $option \leftarrow False$;

\For {$k=1:k_{max}$}
    \State \{$\boldsymbol{\hat{y}}_{n}(\boldsymbol{\theta})\}_{n=0, \dots, N-1} = ODESolve(\boldsymbol{f_\theta}, \boldsymbol{y}_0, (t_0, t_{N-1}))$;
    \State Evaluate normalised objective function $F_{\boldsymbol{\theta}}$;
    
   \State $P_\theta \leftarrow 0$; 
    \For {$i \in \varepsilon$}
        \State Compute penalty term $P_i$ using \eqref{eq:Pi};
        \State Compute self-adaptive penalty parameter $\mu_i$ using \eqref{eq:mu};
        \State $P_{\boldsymbol{\theta}} \leftarrow P_{\boldsymbol{\theta}} + P_i$;
    \EndFor
    \For{$j \in \mathcal{I}$}
        \State Compute penalty term $P_j$ using \eqref{eq:Pj};
        \State Compute self-adaptive penalty parameter $\mu_j$ using \eqref{eq:mu};
        \State $P_{\boldsymbol{\theta}} \leftarrow P_{\boldsymbol{\theta}} + P_j$;
    \EndFor

    \If{$P_{\boldsymbol{\theta}} \leq tol$}
        \State $\phi \leftarrow F_{\boldsymbol{\theta}}$;
    \Else
        \State $\phi \leftarrow F_{\boldsymbol{\theta}} + \dfrac{1}{\#\{\varepsilon\}} \sum_{i \in \varepsilon} \mu_i P_i + \dfrac{1}{\#\{\mathcal{I}\}} \sum_{j \in \mathcal{I}} \mu_j P_j$;
    \EndIf 
    \If{$option == True$}
        \State $\boldsymbol{\theta}_\text{best}, F_\text{best}, P_\text{best} \leftarrow BestPointStrategy(\boldsymbol{\theta}, \boldsymbol{\theta}_\text{best}, F_\text{best}, P_\text{best}, F_{\boldsymbol{\theta}}, P_{\boldsymbol{\theta}})$
        \State $\boldsymbol{\theta} \leftarrow \boldsymbol{\theta}_\text{best}$
    \EndIf
    

    \State $\nabla \phi \leftarrow Optimiser.BackpropCall(\phi)$;
    \State $\boldsymbol{\theta} \leftarrow Optimiser.Step(\nabla \phi, \boldsymbol{\theta})$;
\EndFor

\State \textbf{return} $\boldsymbol{\theta}$;

\end{algorithmic}
\end{algorithm}

\section{Numerical Experiments} \label{sec:experiments}

To assess the effectiveness of using the self-adaptive penalty algorithm for Neural ODEs, three datasets describing natural systems with prior knowledge constraints were used. Namely, the World Population Growth (WPG) \cite{coelho_population_2023}, Chemical Reaction (CR) \cite{coelho_chemical_2023}, and the newly developed Damped Harmonic Oscillator (DHO) \cite{coelho_oscillator_2023}, see Appendix \ref{app:DHO}. Three distinct experiments were carried out to test and validate the models trained using the proposed approach: 

\begin{itemize}
    \item \textbf{reconstruction:} where the same dataset was used for both training and testing. The objective was to assess the model's ability to accurately reproduce the patterns. By using the same dataset for testing, it allowed for a direct comparison between the original data and the reconstructed data generated

    \item \textbf{extrapolation:} where the model was tested on a larger time interval. At this experiment the trained models were put to the test by extrapolating their predictions to a larger time interval than what was used during the training phase. This task aimed to assess the models' generalisation capabilities beyond the temporal scope of the original dataset. By evaluating the accuracy and stability of predictions in this extended time horizon, we could determine the models' robustness in forecasting data beyond their training time interval.

    \item \textbf{completion:} where the test set had a different sampling frequency in the same time interval. This experiment closely resembles real-world scenarios where data is irregularly or sparsely sampled. By assessing the models' ability to handle varying data densities and still provide accurate predictions, we could ascertain their adaptability to practical data collection settings.
\end{itemize}

As a form of comparison, we consider a \emph{vanilla} Neural ODE (a traditional Neural ODE without incorporating constraints) and a Neural ODE with a L1 exact penalty function, using three different $\mu$ values ($\mu=1, 10, 100$) to demonstrate the influence of $\mu$ on the model's performance, which serve as our baselines. Furthermore to study the relevance of the implementation of the \emph{best point} strategy, we present results for the self-adaptive penalty algorithm with and without it.
The assessment of the models was based on the average MSE, $MSE_{avg}$, and average total constraints violation, $P_{avg}$.

\textbf{Remark:} The selected datasets were chosen because they contained valuable information regarding the constraints of the system, which is typically not readily available.

Tables \ref{tab:pop_1}-\ref{tab:visc_4} show the values of $MSE_{avg}$ and $P_{avg}$, and their respective standard deviation (std), after three independent runs, using the self-adaptive penalty algorithm, with and without the \emph{best point} strategy, and two baselines, \emph{vanilla} Neural ODE and a Neural ODE with L1 exact penalty function also with and without the \emph{best point} strategy. 
Figures \ref{fig:pop}-\ref{fig:damped} show the predicted and real curves for the best run at the extrapolation task. Only the plots of the prediction curves for the extrapolation task are displayed, as this task involves forecasting over time horizons not seen during training, thereby allowing a more accurate assessment of whether the constraints were effectively learned.

The code to replicate the experiments presented in this work is available at [LINK]\footnote{available after acceptance}.

\textbf{Remark:} The definition of the inequality constraints is given by the $\max$ function which is non-differentiable in all its domain. Consequently, employing optimisation algorithms reliant on derivatives becomes impractical. However, in deep learning practise, owing to the prevalence of the $\max$ function facilitated by Rectified Linear Units (ReLU), a workaround has been devised. This workaround defines the derivative as $ReLU'(0)=0$, enabling the usage of popular optimisers such as \emph{Adam} avoiding these issues.  As a result, mainstream deep learning libraries like \emph{PyTorch} and \emph{TensorFlow} have incorporated this mechanism \cite{bertoin2021numerical}.

\subsection{World Population Growth}

The WPG dataset is publicly available on \emph{Kaggle} \cite{coelho_population_2023}. It is a time series composed of two features (time step $t_n$, population $y_n$) which describes a system subjected to an inequality constraint defined by the carrying capacity at each time step,

$$\{y_n \leq 12\}, \,\,\, n=0,\dots,N-1.$$

Ideally, all data points of the adjusted model must satisfy this constraint.
In order to evaluate the performance of the NNs in modelling this system.

The trained NNs have $4$ hidden layers: linear with $50$ neurons; hyperbolic tangent (tanh); linear with $50$ neurons; Exponential Linear Unit (ELU). The input and output layers have $1$ neuron. The Adam optimiser was used with a learning rate of $1e-5$. Training was done for $10000$ iterations ($k_{\max}$).
The reconstruction task uses the same set of $200$ data points in the time interval $[0,300]$ for both training and testing. In the extrapolation task, the training set comprises $200$ data points in the time interval $[0,300]$, while the testing set consists of $200$ data points in the interval $[0,400]$. The completion task involves $200$ training points and $300$ testing points in the interval $[0,300]$.

From Tables \ref{tab:pop_1}-\ref{tab:pop_4}, as expected, \emph{vanilla} Neural ODEs show higher $P_{avg}$ and $MSE_{avg}$ values than the L1 exact penalty function ($\mu=1$), with and without the \emph{best point} strategy, and self-adaptive versions. Hence, incorporating constraints into Neural ODEs not only ensures that they are satisfied, but also contributes to better data fitting.

In general, in Neural ODEs with a L1 exact penalty function, with and without the \emph{best point} strategy, a larger $\mu$ results in models with smaller constraints violation value $P_{avg}$ but with a higher $MSE_{avg}$. Inversely, a smaller value of $\mu$ produces models with lower $MSE_{avg}$ values but a higher $P_{avg}$ value.

Neural ODEs with self-adaptive penalty function exhibit the best overall performance in all experiments, with the smallest $MSE_{avg}$ and $P_{avg}$ values. 
Thus, effectively balancing the minimisation of the loss function (fit to the training data) and the constraints violation.
Furthermore, the best performance was achieved without using the \emph{best point} strategy.

In general, the usage of the \emph{best point} strategy tends to result in lower performance, especially when using a L1 exact penalty function, indicating that in this case controlling the optimisation process may not be the optimal strategy, as it restricts exploration.

\begin{figure}[ht]
\begin{center}
\subfigure[]{\includegraphics[width=0.32\linewidth]{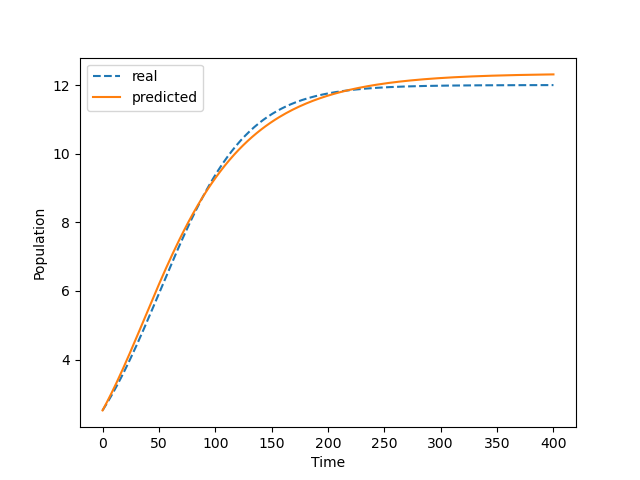}} 
\subfigure[]{\includegraphics[width=0.32\linewidth]{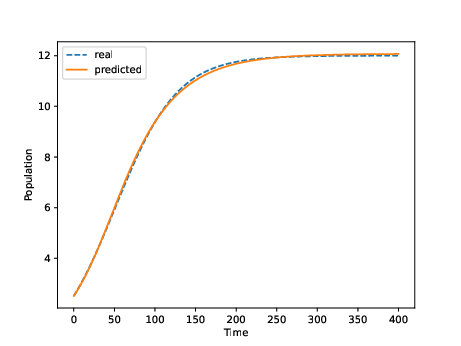}}
\subfigure[]{\includegraphics[width=0.32\linewidth]{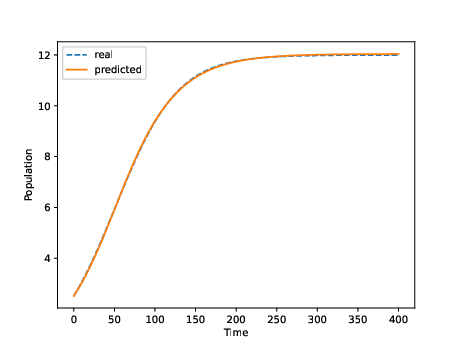}} \\
\subfigure[]{\includegraphics[width=0.32\linewidth]{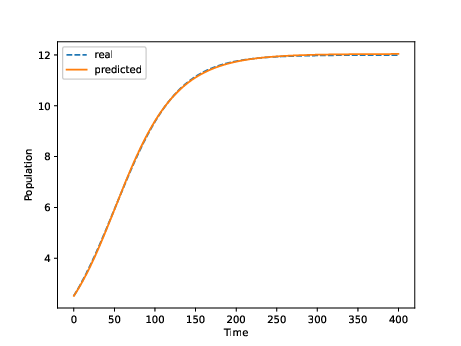}} 
\subfigure[]{\includegraphics[width=0.32\linewidth]{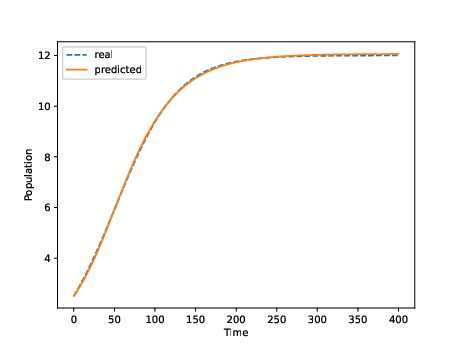}}
\end{center}
\caption{Plot of the real (dashed line) and predicted (solid line) curves, for the WPG dataset, at the extrapolation task trained with \emph{vanilla} Neural ODE (a), Neural ODE with a L1 exact penalty function and $\mu=1$ (b), Neural ODE with a L1 exact penalty function + \emph{best point} and $\mu=1$ (c), self-adaptive penalty function for Neural ODE (d) and self-adaptive penalty function + \emph{best point} for Neural ODE (e).}
\label{fig:pop}
\end{figure}

Figure \ref{fig:pop} displays the predicted and real curves of the testing set for the extrapolation task's best run, chosen due to being the hardest since the models are making predictions beyond the training interval.
Figure \ref{fig:pop}-(a) shows the results for the \emph{vanilla} Neural ODE, which was unable to capture the inequality constraint from the training data. This is evident from the exponential increase in population values over time. In contrast, the Neural ODEs with a L1 exact penalty function and self-adaptive penalty function, with and without \emph{best point} strategy, closely fit the real curve and demonstrates their modelling superiority.

These results showcase the difficulty on selecting an appropriate value for $\mu$ of L1 penalty function. Thus, the proposed self-adaptive penalty function overcomes this issue and produces the best results overall, being competitive.

\begin{table}[ht]
\caption{Performance on WPG dataset of \emph{vanilla} Neural ODE.}
\label{tab:pop_1}
\begin{tabular}{@{}ccc@{}}
\toprule
               & \multicolumn{2}{c}{\multirow{2}{*}{Vanilla Neural ODE}} \\
               & \multicolumn{2}{c}{}           \\ \midrule
Experiment     & $\text{$MSE_{avg}$} \pm \text{std}$    & $P_{avg} \pm \text{std}$     \\ \midrule
Reconstruction & 1.9e-1 $\pm$ 1.5e-1            & 7.8e-2 $\pm$ 5.1e-2    \\
Extrapolation  & 1.3e-1 $\pm$ 7.2e-2            & 1.5e-1 $\pm$ 5.3e-2    \\
Completion     & 4.3e-2 $\pm$ 2.9e-3            & 2.8e-2 $\pm$ 1.2e-2    \\ \bottomrule
\end{tabular}
\end{table}

\begin{table}[ht]
\caption{Performance on WPG dataset of Neural ODE with a L1 exact penalty function with the \emph{best point} strategy ($\mu=1, 10, 100$).}
\label{tab:pop_2}
\tabcolsep=0.10cm
\begin{tabular}{@{}ccccccc@{}}
\toprule
               & \multicolumn{6}{c}{L1 exact penalty function + \emph{best point}}                                                                                                          \\ \cmidrule(l){2-7} 
               & \multicolumn{2}{c}{$\mu=1$}                      & \multicolumn{2}{c}{$\mu=10$}                     & \multicolumn{2}{c}{$\mu=100$}                    \\ \midrule
Experiment     & $\text{$MSE_{avg}$} \pm \text{std}$ & $P_{avg} \pm \text{std}$ & $\text{$MSE_{avg}$} \pm \text{std}$ & $P_{avg} \pm \text{std}$ & $\text{$MSE_{avg}$} \pm \text{std}$ & $P_{avg} \pm \text{std}$ \\ \midrule
Reconstruction &      4.9e-2 $\pm$ 4.0e-2                       &    5.5e-3 $\pm$ 4.8e-3                &      3.5e-2 $\pm$ 2.4e-2                       &    1.9e-4 $\pm$ 1.8e-4                &     1.7e-1 $\pm$ 1.3e-1                        &  3.1e-3 $\pm$ 1.8e-3                  \\
Extrapolation  &     1.3e-1 $\pm$ 9.0e-2                        &    9.1e-2 $\pm$ 6.0e-2                &        1.1e1 $\pm$ 8.0                     &  2.7 $\pm$ 2.3                  &     1.3 $\pm$ 6.3e-1                        &  1.1 $\pm$ 3.5e-1                  \\
Completion     &      3.8e-2 $\pm$ 1.0e-2                       &   3.9e-3 $\pm$ 1.1e-3                 &         4.5e-2 $\pm$ 3.0e-2                    &    5.1e-3 $\pm$ 6.9e-3                &     8.0e-2 $\pm$ 1.6e-2                        &    4.3e-3 $\pm$ 1.9e-3                \\ \bottomrule
\end{tabular}
\end{table}

\begin{table}[ht]
\caption{Performance on WPG dataset of Neural ODE with a L1 exact penalty function ($\mu=1, 10, 100$).}
\label{tab:pop_3}
\tabcolsep=0.10cm
\begin{tabular}{@{}ccccccc@{}}
\toprule
               & \multicolumn{6}{c}{L1 exact penalty function}                                                                                                                   \\ \cmidrule(l){2-7} 
               & \multicolumn{2}{c}{$\mu=1$}                         & \multicolumn{2}{c}{$\mu=10$}                        & \multicolumn{2}{c}{$\mu=100$}                       \\ \midrule
Experiment     & $\text{$MSE_{avg}$} \pm \text{std}$ & $P_{avg} \pm \text{std}$    & $\text{$MSE_{avg}$} \pm \text{std}$ & $P_{avg} \pm \text{std}$    & $\text{$MSE_{avg}$} \pm \text{std}$ & $P_{avg} \pm \text{std}$    \\ \midrule
Reconstruction & 3.3e-2 $\pm$ 2.5e-2       & 3.3e-3 $\pm$ 2.4e-3 & 2.8e-2 $\pm$ 1.4e-3       & 2.3e-5 $\pm$ 1.7e-5 & 3.4e-1 $\pm$ 3.1e-1       & 2.9e-5 $\pm$ 3.5e-5 \\
Extrapolation  & 4.9e-2 $\pm$ 5.9e-2       & 4.8e-2 $\pm$ 4.5e-2 & 6.4e-2 $\pm$ 5.1e-2       & 2.5e-2 $\pm$ 1.5e-2 & 6.4e-2 $\pm$ 2.9e-2       & 1.6e-2 $\pm$ 8.2e-3 \\
Completion     & 1.5e-2 $\pm$ 5.3e-3       & 1.4e-3 $\pm$ 5.4e-4 & 2.7e-2 $\pm$ 1.9e-2       & 2.5e-5 $\pm$ 2.4e-5 & 1.3e-1 $\pm$ 7.2e-3       & 1.5e-5 $\pm$ 2.1e-5 \\ \bottomrule
\end{tabular}%
\end{table}

\begin{table}[ht]
\caption{Performance on WPG dataset of self-adaptive penalty function for Neural ODE.}
\label{tab:pop_4}
\begin{tabular}{@{}ccccc@{}}
\toprule
               & \multicolumn{2}{c}{\multirow{2}{*}{\begin{tabular}[c]{@{}c@{}}Self-adaptive Penalty Function\\ + \emph{best point}\end{tabular}}} & \multicolumn{2}{c}{\multirow{2}{*}{Self-adaptive Penalty Function}} \\
               & \multicolumn{2}{c}{}                                                     & \multicolumn{2}{c}{}                                                 \\ \midrule
Experiment     & $\text{$MSE_{avg}$} \pm \text{std}$                                              & $P_{avg} \pm \text{std}$                                              & $\text{$MSE_{avg}$} \pm \text{std}$ & \multicolumn{1}{c}{$P_{avg} \pm \text{std}$} \\ \midrule
Reconstruction &            1.1e-3 $\pm$ 2.6e-4                                                              &       4.9e-4 $\pm$ 8.5e-5                                                           &               8.5e-4 $\pm$ 9.3e-5              &        4.2e-4 $\pm$ 3.4e-5                                \\
Extrapolation  &         1.6e-3 $\pm$ 3.4e-4                                                                 &   1.1e-2 $\pm$ 1.3e-3                                                              &            1.3e-3 $\pm$ 4.2e-4                 &      9.4e-3 $\pm$ 1.4e-6                                  \\
Completion     &        1.4e-3 $\pm$ 3.6e-4                                                                  &    7.8e-4 $\pm$ 3.2e-4                                                             &        1.0e-3 $\pm$ 2.3e-4                     &      4.8e-4 $\pm$ 1.3e-4                                  \\ \bottomrule
\end{tabular}
\end{table}

\subsection{Chemical Reaction}

The CR dataset is publicly available on \emph{Kaggle}, which describes the evolution of the species in a chemical reaction \cite{coelho_chemical_2023}. It is a time-series composed of five features (time step $t_n$, and the masses of species A, $y_A$, B, $y_B$, C, $y_C$, and D, $y_D$ with $\boldsymbol{y}_n=(y_A,y_B,y_C,y_D)$). At each time step $t_n$ ($n=0,\dots,N-1$), this system has an equality constraint defined by the conservation of mass: 

$$\boldsymbol{y}_n = \{y_A + y_B + y_C + y_D\}_n = m_{\text{total}}, \,\,\ n=0,\dots,N-1$$

\noindent with $m_{\text{total}}>0$ the total constant mass of the system. Ideally, all data points of the adjusted model must satisfy this constraint.

The NNs were trained for $10000$ iterations ($k_{\max}$) and have $6$ hidden layers: linear with $50$ neurons; tanh; linear with $64$ neurons; ELU; linear with $50$ neurons; tanh. The input and output layers have $4$ neurons. The Adam optimiser was used with a learning rate of $1e-5$.
For the reconstruction task, both the training and testing sets consist of $100$ data points in the time interval $[0,100]$. The extrapolation task involves training with $100$ data points in the time interval $[0,100]$ and testing with $100$ data points in the interval $[0,200]$. The completion task uses $100$ training points and $200$ testing points in the interval $[0,100]$.


Tables \ref{tab:react_1}-\ref{tab:react_4} present the results obtained for the CR dataset. In general, from Tables \ref{tab:react_1}-\ref{tab:react_4}, it is observed that the performance of the \emph{vanilla} Neural ODE surpasses that of the Neural ODE with an L1 exact penalty function with the \emph{best point} strategy, while being comparable without it, for $\mu=1$. Increasing $\mu$ to $10$ and $100$ improves the violation of the constraints, except for the extrapolation task with $\mu=100$.

Optimal performance, lower $P_{avg}$ and $MSE_{avg}$ values, was achieved with the proposed self-adaptive penalty function, exhibiting no significant difference between with or without the \emph{best point} strategy.

\begin{figure}[ht]
\begin{center}
\subfigure[]{\includegraphics[width=0.32\linewidth]{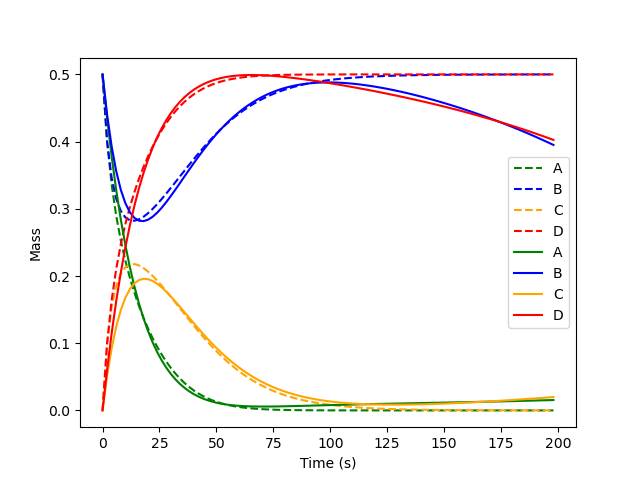}} 
\subfigure[]{\includegraphics[width=0.32\linewidth]{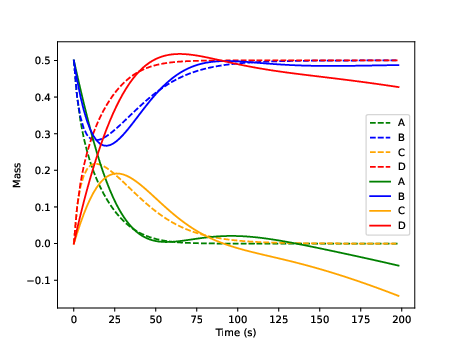}}
\subfigure[]{\includegraphics[width=0.32\linewidth]{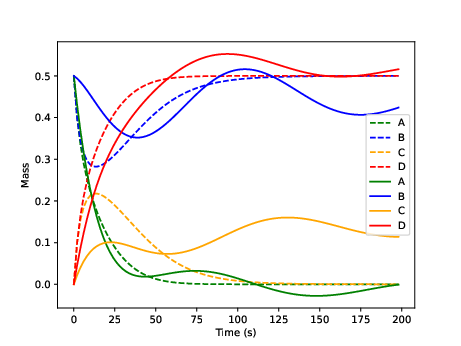}} \\
\subfigure[]{\includegraphics[width=0.32\linewidth]{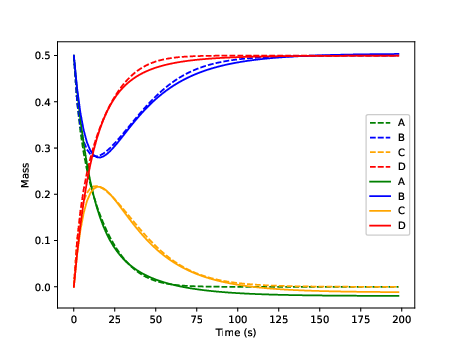}} 
\subfigure[]{\includegraphics[width=0.32\linewidth]{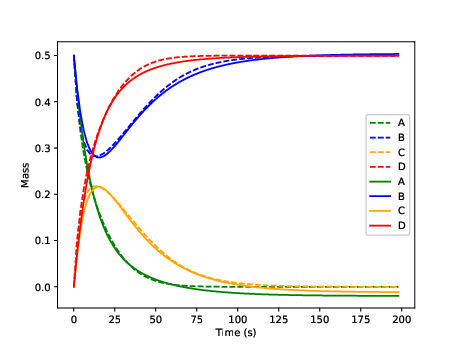}}
\end{center}
\caption{Plot of the real (dashed line) and predicted (solid line) curves, for the CR dataset, at the extrapolation task trained with \emph{vanilla} Neural ODE (a), Neural ODE with a L1 exact penalty function and $\mu=1$ (b), Neural ODE with a L1 exact penalty function + \emph{best point} and $\mu=1$ (c), self-adaptive penalty function for Neural ODE (d) and self-adaptive penalty function + \emph{best point} for Neural ODE (e).}
\label{fig:react}
\end{figure}

Figure \ref{fig:react} displays the predicted and real curves of the testing set for the extrapolation task's best run for all Neural ODE baselines and self-adaptive penalty function.
As it can be seen, the Neural ODE with the self-adaptive penalty function with, \ref{fig:react}-(d), and without \ref{fig:react}-(e), the \emph{best point} strategy, effectively and accurately model the data dynamics. This evidences that the equality constraint was successfully learnt during training. This contrasts with \emph{vanilla} Neural ODE, \ref{fig:react}-(a), and Neural ODEs with the L1 exact penalty function, \ref{fig:react}-(b)(c), with and without the \emph{best point} strategy, which clearly could not capture the dynamics of the data, deviating from the real curves.



\begin{table}[ht]
\caption{Performance on CR dataset of \emph{vanilla} Neural ODE.}
\label{tab:react_1}
\begin{tabular}{@{}ccc@{}}
\toprule
               & \multicolumn{2}{c}{\multirow{2}{*}{Vanilla Neural ODE}} \\
               & \multicolumn{2}{c}{}           \\ \cmidrule(l){2-3} 
Experiment     & $\text{$MSE_{avg}$} \pm \text{std}$    & $P_{avg} \pm \text{std}$     \\ \midrule
Reconstruction & 2.1e-3 $\pm$ 1.4e-3            & 7.8e-3 $\pm$ 6.7e-3    \\
Extrapolation  & 4.1e-2 $\pm$ 5.7e-2            & 1.6e-2 $\pm$ 1.6e-2    \\
Completion     & 9e-4 $\pm$ 5.9e-4     & 2.6e-3 $\pm$ 3.1e-3    \\ \bottomrule
\end{tabular}
\end{table}

\begin{table}[ht]
\tabcolsep=0.10cm
\caption{Performance on CR dataset of Neural ODE with a L1 exact penalty function with the \emph{best point} strategy ($\mu=1, 10, 100$).}
\label{tab:react_2}
\begin{tabular}{@{}ccccccc@{}}
\toprule
               & \multicolumn{6}{c}{L1 exact penalty function + \emph{best point}}                                                                                                                   \\ \cmidrule(l){2-7} 
               & \multicolumn{2}{c}{$\mu=1$}                         & \multicolumn{2}{c}{$\mu=10$}                        & \multicolumn{2}{c}{$\mu=100$}                       \\ \midrule
Experiment     & $\text{$MSE_{avg}$} \pm \text{std}$ & $P_{avg} \pm \text{std}$    & $\text{$MSE_{avg}$} \pm \text{std}$ & $P_{avg} \pm \text{std}$    & $\text{$MSE_{avg}$} \pm \text{std}$ & $P_{avg} \pm \text{std}$    \\ \midrule
Reconstruction & 1.1e-1 $\pm$ 1.2e-1       & 4.8e-2 $\pm$ 5.3e-2 & 3.5e-2 $\pm$ 2.4e-2       & 1.9e-4 $\pm$ 1.8e-4 & 1.7e-1 $\pm$ 1.3e-1       & 3.1e-3 $\pm$ 1.8e-3 \\
Extrapolation  & 4.6e-2 $\pm$ 5.7e-2       & 2.9e-1 $\pm$ 3.7e-1 & 1.1e1 $\pm$ 8.0       & 2.7 $\pm$ 2.3 & 1.3 $\pm$ 6.3e-1       & 1.1 $\pm$ 3.5e-1 \\
Completion     & 8.0e-2 $\pm$ 1.1e-1       & 1.2e-1 $\pm$ 1.6e-0  & 4.5e-2 $\pm$ 3.0e-2       & 5.1e-3 $\pm$ 6.9e-3 & 8.0e-2 $\pm$ 1.6e-2       & 4.3e-3 $\pm$ 1.9e-3 \\ \bottomrule
\end{tabular}%
\end{table}

\begin{table}[ht]
\caption{Performance on CR dataset of Neural ODE with a L1 exact penalty function ($\mu=1, 10, 100$).}
\label{tab:react_3}
\tabcolsep=0.10cm
\begin{tabular}{@{}lcccccc@{}}
\toprule
               & \multicolumn{6}{c}{L1 exact penalty function}                                                                                                                   \\ \cmidrule(l){2-7} 
               & \multicolumn{2}{c}{$\mu=1$}                         & \multicolumn{2}{c}{$\mu=10$}                        & \multicolumn{2}{c}{$\mu=100$}                       \\ \midrule
Experiment     & $\text{$MSE_{avg}$} \pm \text{std}$ & $P_{avg} \pm \text{std}$    & $\text{$MSE_{avg}$} \pm \text{std}$ & $P_{avg} \pm \text{std}$    & $\text{$MSE_{avg}$} \pm \text{std}$ & $P_{avg} \pm \text{std}$    \\ \midrule
Reconstruction & 6.6e-3 $\pm$ 3.4e-3       & 2.3e-3 $\pm$ 1.5e-3 & 1.8e-2 $\pm$ 1.0e-2       & 6.0e-3 $\pm$ 4.5e-3 & 4.1e-1 $\pm$ 5.2e-1       & 5.9e-3 $\pm$ 3.0e-3 \\
Extrapolation  & 3.7e-2 $\pm$ 3.7e-2       & 2.5e-1 $\pm$ 1.8e-1 & 2.0e-1 $\pm$ 1.2e-1       & 1.8e-1 $\pm$ 1.9e-1 & 3.3 $\pm$ 2.2       & 1.8 $\pm$ 2.5e-1 \\
Completion     & 9.6e-3 $\pm$ 1.2e-3       & 9.2e-3 $\pm$ 9.3e-3 & 1.7e-2 $\pm$ 1.3e-2       & 3.0e-3 $\pm$ 2.0e-3 & 2.8e-1 $\pm$ 2.9e-1       & 5.8e-3 $\pm$ 5.1e-3 \\ \bottomrule
\end{tabular}%
\end{table}

\begin{table}[ht]
\caption{Performance on CR dataset of self-adaptive penalty function for Neural ODE.}
\label{tab:react_4}
\begin{tabular}{@{}lcccl@{}}
\toprule
               & \multicolumn{2}{c}{\multirow{2}{*}{\begin{tabular}[c]{@{}c@{}}Self-adaptive Penalty Function\\ + \emph{best point}\end{tabular}}} & \multicolumn{2}{c}{\multirow{2}{*}{Self-adaptive Penalty Function}} \\
               & \multicolumn{2}{c}{}                                                     & \multicolumn{2}{c}{}                                                 \\ \midrule
Experiment     & $\text{$MSE_{avg}$} \pm \text{std}$                                              & $P_{avg} \pm \text{std}$                                              & $\text{$MSE_{avg}$} \pm \text{std}$ & \multicolumn{1}{c}{$P_{avg} \pm \text{std}$} \\ \midrule
Reconstruction &      1.1e-4 $\pm$ 9.6e-5                                                                    &                    1.0e-6 $\pm$ 1.0e-6                                             &                 1.4e-4 $\pm$ 9.6e-5            &   1.0e-6 $\pm$ 1.0e-6                                     \\
Extrapolation  &        7.3e-5 $\pm$ 2.0e-5                                                                  &                      1.0e-6 $\pm$ 1.0e-6                                           &               7.9e-5 $\pm$ 2.0e-5              &    1.0e-6 $\pm$ 1.0e-6                                    \\
Completion     &      5.6e-4 $\pm$ 7.0e-4                                                                    &                       3.7e-5 $\pm$ 4.1e-5                                          &            1.6e-3 $\pm$ 1.0e-6                 &    9.3e-5 $\pm$ 1.0e-6                                    \\ \bottomrule
\end{tabular}
\end{table}

\subsection{Damped Harmonic Oscillator}


The DHO dataset, available on \emph{Kaggle}, describes the motion of a mass attached to a spring under the influence of a dissipative force described by a second-order ODE \cite{coelho_oscillator_2023}. The dataset consists of a time series with four features: time step $t_n$, displacement $y_x$, velocity $y_v$, and acceleration $y_a$ ($\boldsymbol{y}_n=(y_x,y_v,y_a)$). 
At each time step $t_n$, the system has one inequality constraint and one equality constraint. The inequality constraint requires the system's energy to decrease due to damping: 

$$E_{t_n} > E_{t_{n+1}}, \,\,\, n=0,\dots,N-1,$$

\noindent where $E_{t_n} = \{\dfrac{1}{2} m\ y_v^2 + \dfrac{1}{2} k y_x^2\}_n$ with $n=0,\dots,N-1$, and $k$ is the spring constant. 

The equality constraint requires the conservation of the rate at which energy is dissipated due to damping: 

$$\Delta P_{t_n} = 0, \,\,\, n=0,\dots,N-1,$$

\noindent where $P_{t_n} = \{-c y_v y_x\}_n$, and $c$ is the damping coefficient. To obtain meaningful results, all data points of the adjusted model should satisfy these constraints.



The DHO dataset is unique among the datasets used in this study because it involves a system with two constraints. 



The trained NNs have $4$ hidden layers: linear with $50$ neurons; hyperbolic tangent (tanh); linear with $50$ neurons; Exponential Linear Unit (ELU). The input and output layers have $2$ neurons. The Adam optimiser was used with a learning rate of $1e-5$ and the optimisation process was done until $10000$ iterations ($k_{\max}$) were done.
In the reconstruction task, the training and testing sets contain the same $400$ data points in the time interval $[0,50]$. The extrapolation task involves training with $400$ data points in the time interval $[0,50]$ and testing with $400$ data points in the interval $[0,400]$. For the completion task, there are $400$ training points and $600$ testing points in the interval $[0,50]$.

The numerical results presented in Tables \ref{tab:visc_1}-\ref{tab:visc_4}  show that Neural ODEs using a L1 exact penalty function with the \emph{best point} strategy shows the worst performance, failing completely at the extrapolation task. However, when compared to a \emph{vanilla} Neural ODE, constraint violation values were similar or lower for the reconstruction and completion tasks when the \emph{best point} strategy was not used.

In contrast, the Neural ODEs with the self-adaptive penalty function and with and without \emph{best point} strategy, performed consistently better, exhibiting a notable difference in magnitude for both $MSE_{avg}$ and constraint violation $P_{avg}$ when compared to the baselines. Moreover, the standard deviation values indicate that this method produced the most stable models across all three runs. Furthermore, the usage of the \emph{best point} strategy slightly improves the $MSE_{avg}$ values.

\begin{figure}[ht]
\begin{center}
\subfigure[]{\includegraphics[width=0.32\linewidth]{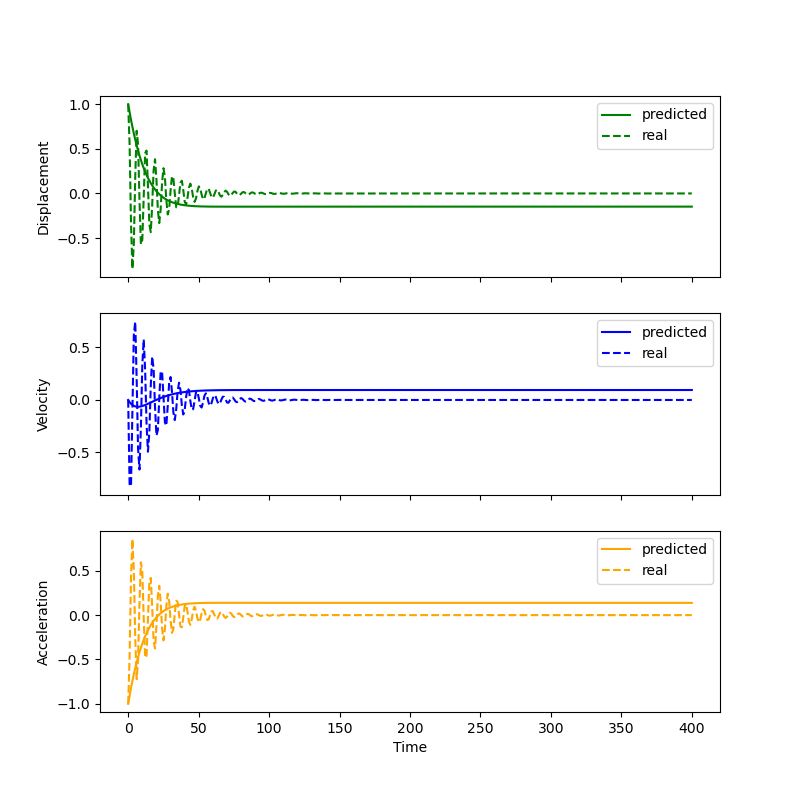}} 
\subfigure[]{\includegraphics[width=0.32\linewidth]{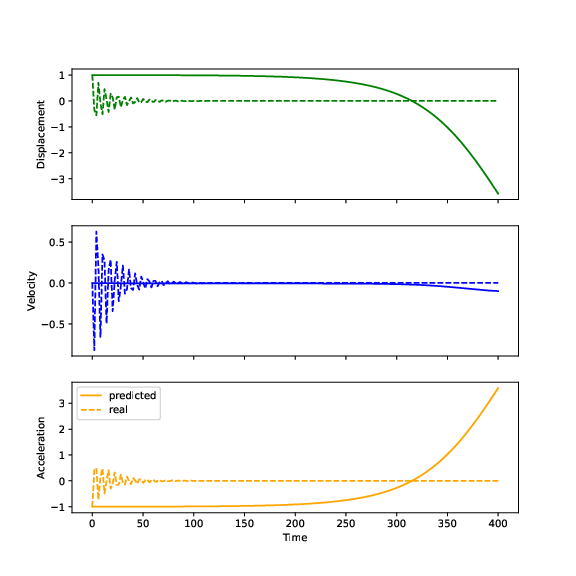}}
\subfigure[]{\includegraphics[width=0.32\linewidth]{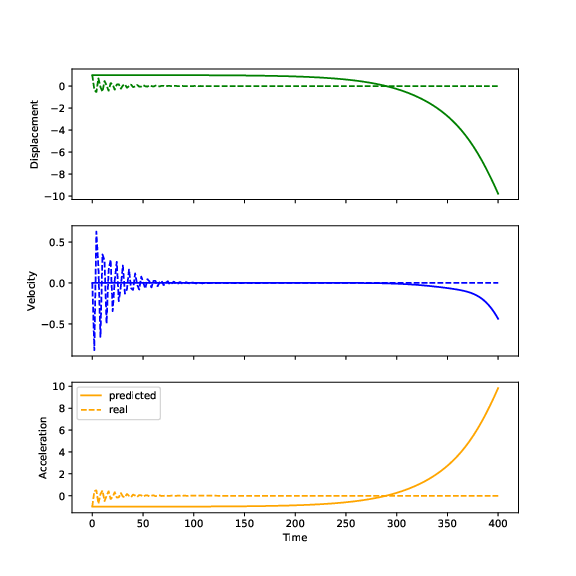}} \\
\subfigure[]{\includegraphics[width=0.32\linewidth]{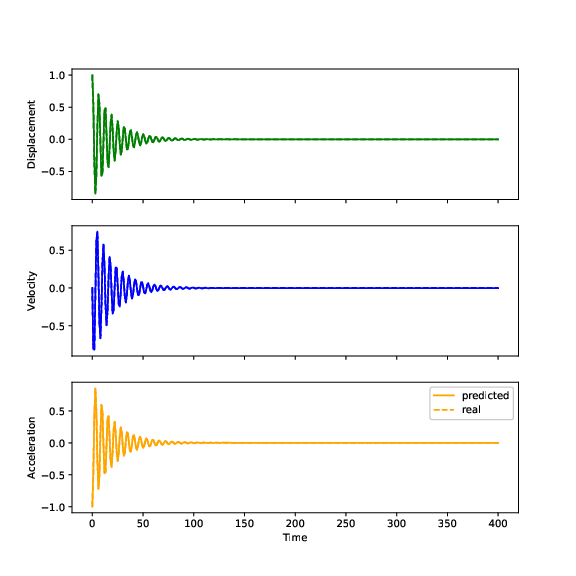}} 
\subfigure[]{\includegraphics[width=0.32\linewidth]{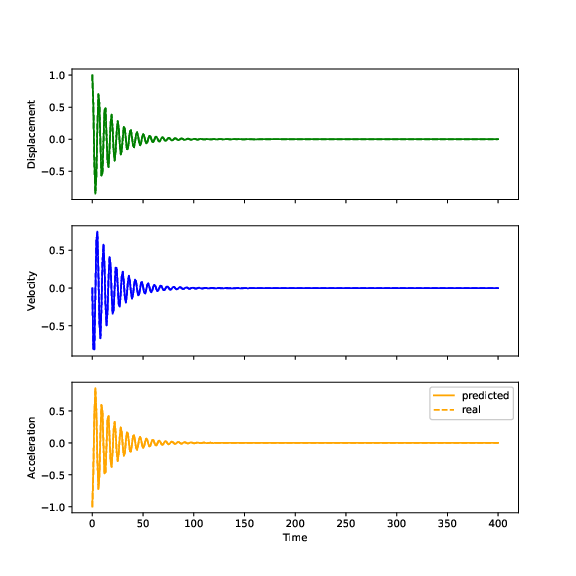}}
\end{center}
\caption{Plot of the real (dashed line) and predicted (solid line) curves, for the DHO dataset, at the extrapolation task trained with \emph{vanilla} Neural ODE (a), Neural ODE with a L1 exact penalty function and $\mu=1$ (b), Neural ODE with a L1 exact penalty function + \emph{best point} and $\mu=1$ (c), self-adaptive penalty function for Neural ODE (d) and self-adaptive penalty function + \emph{best point} for Neural ODE (e).}
\label{fig:damped}
\end{figure}

Figure \ref{fig:damped} displays the predicted and real curves of the testing set for the extrapolation task's best run for all Neural ODE baselines and self-adaptive penalty function.
Figure \ref{fig:damped} shows both \emph{vanilla} Neural ODE and the Neural ODE with a L1 exact penalty function, with and without the \emph{best point} strategy, struggled to model this system, with predicted curves largely deviating from the real curves. In contrast, the Neural ODE with the self-adaptive penalty function, with and without the \emph{best point} strategy, accurately adjusted to the data dynamics. This shows that the self-adaptive penalty function allows Neural ODEs to effectively incorporate the constraints into the model, as demonstrated by the model's ability to capture the physical dynamics of the system even when predicting for longer time horizons not seen during training. This behaviour was not observed with either the \emph{vanilla} Neural ODE or the Neural ODE with a L1 exact penalty function, demonstrating the superiority of our proposed self-adaptive penalty function.

\begin{table}[ht]
\caption{Performance on DHO dataset of \emph{vanilla} Neural ODE.}
\label{tab:visc_1}
\begin{tabular}{@{}ccc@{}}
\toprule
               & \multicolumn{2}{c}{\multirow{2}{*}{Vanilla Neural ODE}} \\
               & \multicolumn{2}{c}{}           \\ \cmidrule(l){2-3} 
Experiment     & $\text{$MSE_{avg}$} \pm \text{std}$    & $P_{avg} \pm \text{std}$     \\ \midrule
Reconstruction & 1.2e-1 $\pm$ 2.9e-2            & 1.2e-3 $\pm$ 1.5e-3    \\
Extrapolation  & 8.5 $\pm$ 11.9                 & 13.9 $\pm$ 19.7        \\
Completion     & 1e-1 $\pm$ 9.3e-3              & 1.4e-4 $\pm$ 6e-5      \\ \bottomrule
\end{tabular}
\end{table}

\begin{table}[ht]
\caption{Performance on DHO dataset of Neural ODE with a L1 exact penalty function  with the \emph{best point} strategy ($\mu=1, 10, 100$).}
\label{tab:visc_2}
\tabcolsep=0.10cm
\begin{tabular}{@{}ccccccc@{}}
\toprule
               & \multicolumn{6}{c}{L1 exact penalty function + \emph{best point}}                                                                                                          \\ \cmidrule(l){2-7} 
               & \multicolumn{2}{c}{$\mu=1$}                      & \multicolumn{2}{c}{$\mu=10$}                     & \multicolumn{2}{c}{$\mu=100$}                    \\ \midrule
Experiment     & $\text{$MSE_{avg}$} \pm \text{std}$ & $P_{avg} \pm \text{std}$ & $\text{$MSE_{avg}$} \pm \text{std}$ & $P_{avg} \pm \text{std}$ & $\text{$MSE_{avg}$} \pm \text{std}$ & $P_{avg} \pm \text{std}$ \\ \midrule
Reconstruction &       3.0e-1 $\pm$ 3.4e-1                      &   5.3e-3 $\pm$ 1.7e-3                 &     7.2e+2 $\pm$ 1.0e+3                        &    8.3e-1 $\pm$ 1.1                &    7.2 $\pm$ 8.4                         &     3.0e-3 $\pm$ 3.1e-3               \\
Extrapolation  &      3.3 $\pm$ 2.5                       &    1.6e-1 $\pm$ 1.8e-1                &     2.3 $\pm$ 1.4                        &  2.7e-1 $\pm$ 3.3e-1                  &             1.1 $\pm$ 1.0                &     2.0e-2 $\pm$ 2.7e-2               \\
Completion     &     4.2e-2 $\pm$ 4.6e-2                        &    2.9e-3 $\pm$ 2.6e-3                &          1.1 $\pm$ 1.1                   &    2.4e-2 $\pm$ 3.3e-2                &    3.3e+1 $\pm$ 4.5e+1                         &    3.2e-2 $\pm$ 2.3e-2                \\ \bottomrule
\end{tabular}
\end{table}

\begin{table}[ht]
\caption{Performance on DHO dataset of Neural ODE with a L1 exact penalty function ($\mu=1, 10, 100$).}
\label{tab:visc_3}
\tabcolsep=0.10cm
\begin{tabular}{@{}lcccccc@{}}
\toprule
               & \multicolumn{6}{c}{L1 exact penalty function}                                                                                                                   \\ \cmidrule(l){2-7} 
               & \multicolumn{2}{c}{$\mu=1$}                         & \multicolumn{2}{c}{$\mu=10$}                        & \multicolumn{2}{c}{$\mu=100$}                       \\ \midrule
Experiment     & $\text{$MSE_{avg}$} \pm \text{std}$ & $P_{avg} \pm \text{std}$    & $\text{$MSE_{avg}$} \pm \text{std}$ & $P_{avg} \pm \text{std}$    & $\text{$MSE_{avg}$} \pm \text{std}$ & $P_{avg} \pm \text{std}$    \\ \midrule
Reconstruction & 1.1e+1 $\pm$ 1.5e+1       & 2.3e-2 $\pm$ 2.9e-2 & 1.5e+1 $\pm$ 2.1e+1       & 6.7e-2 $\pm$ 9.4e-2 & 2.0e-1 $\pm$ 9.0e-2       & 6.0e-4 $\pm$ 7.1e-5 \\
Extrapolation  & 5.3e-1 $\pm$ 1.5e-1       & 3.6e-2 $\pm$ 3.0e-2 & 4.2e-2 $\pm$ 3.9e-2       & 2.0e-4 $\pm$ 2.8e-4 & 2.2e-2 $\pm$ 4.1e-3       & 1.0e-6 $\pm$ 1.0e-6 \\
Completion     & 1.1 $\pm$ 1.3       & 1.1e-2 $\pm$ 1.5e-2 & 2.7e-1 $\pm$ 1.9e-1       & 3.1e-3 $\pm$ 4.4e-3 & 3.4e+2 $\pm$ 4.7e+2       & 6.0e-2 $\pm$ 8.4e-2 \\ \bottomrule
\end{tabular}%
\end{table}

\begin{table}[ht]
\caption{Performance on DHO dataset of self-adaptive penalty function for Neural ODE.}
\label{tab:visc_4}
\begin{tabular}{@{}lcccl@{}}
\toprule
               & \multicolumn{2}{c}{\multirow{2}{*}{\begin{tabular}[c]{@{}c@{}}Self-adaptive Penalty Function\\ + \emph{best point}\end{tabular}}} & \multicolumn{2}{c}{\multirow{2}{*}{Self-adaptive Penalty Function}} \\
               & \multicolumn{2}{c}{}                                                     & \multicolumn{2}{c}{}                                                 \\ \midrule
Experiment     & $\text{$MSE_{avg}$} \pm \text{std}$                                              & $P_{avg} \pm \text{std}$                                              & $\text{$MSE_{avg}$} \pm \text{std}$ & \multicolumn{1}{c}{$P_{avg} \pm \text{std}$} \\ \midrule
Reconstruction &                       1.2e-4 $\pm$ 1.4e-4                                                   &               6.3e-3 $\pm$ 1.9e-5                                                   &         3.3e-5 $\pm$ 2.7e-5                    &        6.3e-3 $\pm$ 9.0e-6                                \\
Extrapolation  &         4.2e-5 $\pm$ 5.1e-5                                                                 &               7.8e-4 $\pm$ 2.0e-6                                                 &        4.0e-6 $\pm$ 1.0e-6                      &     7.8e-4 $\pm$ 1.0e-6                                   \\
Completion     &           5.1e-5 $\pm$ 3.9e-5                                                               &      6.3e-3 $\pm$ 7.0e-6                                                           &        6.8e-5 $\pm$ 7.5e-5                     &       6.3e-3 $\pm$ 7.8e-5                                  \\ \bottomrule
\end{tabular}
\end{table}

\clearpage

\section{Conclusion} \label{sec:conclusion}

This work proposes a self-adaptive penalty function and a self-adaptive penalty algorithm for Neural ODEs to enable modelling of constrained natural systems. The proposed self-adaptive penalty algorithm can dynamically adjust the penalty parameters.

 Our approach is an improvement over traditional penalty methods which require an appropriate initialisation and tuning of penalty parameters $\mu$ during the optimisation process. This selection is challenging and time-consuming. In general, in the context of NNs, a fixed penalty parameter is used, and consequently an optimal solution may not be found.
 
 The proposed self-adaptive penalty algorithm dynamically adjusts penalty parameters taking into account the degree of constraints violation, resulting in more efficient and accurate predictions, especially for complex systems with constraints. The self-adaptive penalty function employs a normalisation step to ensure all values (loss and constraints violations) have the same order of magnitude, improving training stability, critical for complex systems prone to numerical instability.

To evaluate the self-adaptive penalty algorithm for Neural ODEs, we used three constrained natural systems and tested the performance on three tasks: reconstruction, extrapolation, and completion. We compared the results obtained by our algorithm with two baselines, a \emph{vanilla} Neural ODE and three Neural ODEs with L1 exact penalty function, with and without the \emph{best point} strategy, and penalty parameter values $\mu=1,10,100$, respectively. We measured performance in terms of $MSE_{avg}$ and total constraints violation $P_{avg}$. Furthermore, the predicted and real curves at the extrapolation task were plotted and analysed.
Our algorithm produced models that achieved remarkably higher performance, exhibiting lower $MSE_{avg}$ and $P_{avg}$ values, indicating that the models not only fit data but also satisfy the governing laws of constrained systems. Additionally, the models produced with our algorithm showed the ability to extrapolate and generalise well. Moreover, the usage of the \emph{best point} strategy does not show a significant change in their performance, however its usage guarantees a degree of control over the optimisation process.

The proposed self-adaptive penalty function and self-adaptive penalty algorithm for Neural ODEs can be applied to any NN architecture and represents a promising approach for incorporating prior knowledge constraints into NNs while enhancing interpretability. Future work will focus on modelling more complex scenarios with more complex constraints using Neural ODEs with our self-adaptive penalty algorithm, as well as conducting theoretical studies on the convergence of our approach.

\begin{acks}
The authors acknowledge the funding by Fundação para a Ciência e Tecnologia (Portuguese Foundation for Science and Technology) through CMAT projects UIDB/00013/2020 and UIDP/00013/2020 and the funding by FCT and Google Cloud partnership through projects CPCA-IAC/AV/589164/2023 and CPCA-IAC/AF/589140/2023.

\noindent C. Coelho would like to thank FCT the funding through the scholarship with reference 2021.05201.BD.

This work is also financially supported by national funds through the FCT/MCTES (PIDDAC), under the project 2022.06672.PTDC - iMAD - Improving the Modelling of Anomalous Diffusion and Viscoelasticity: solutions to industrial problems.    
\end{acks}

\begin{appendices}

\section{Developed Dataset} \label{app:DHO}

\paragraph{Development of the Damped Harmonic Oscillator Dataset}

The DHO dataset was intentionally created to systematically assess the boundaries of the proposed self-adaptive penalty method. This was motivated by the heightened complexity and non-periodic nature of the dataset's dynamics.

The DHO dataset simulates the behaviour of a damped harmonic oscillator over time. The dynamics of this system follow a decaying oscillatory pattern influenced by the interplay between the restoring force from the spring and the damping force, resulting in gradual attenuation of amplitude over time. The dynamics of the damped harmonic oscillator are governed by the second-order ODE:

\begin{equation}
    \begin{cases}
        m x''(t) + c x'(t) + kx(t) = 0, \\
        x(0) = 1, \\
        x'(0) = 0
    \end{cases}
\end{equation}

\noindent where $x(t)$ is the displacement, $x'(t)$ is the velocity and $x''(x)$ the acceleration, at time $t$, with mass, $m$, of 1kg in a mass-spring-damper system, a spring constant $k$ of 1N/m and a damping coefficient $c$ of 0.1 Ns/m.
The oscillator is initially displaced by 1m and an initial velocity of 0m/s.

This dataset translates a physical system governed by physical laws. Thus, it possesses two prior knowledge constraints that must be satisfied at all time steps. The energy $E(t)$ of the system decreases through time until the mass comes to rest,

$$E(t) > E(t+1),$$

\noindent where $E(t)=\dfrac{1}{2} \,m y_v(t)^2 + \dfrac{1}{2}\, k y_x(t)^2$, with $y_x=x(t)$ is the displacement and $y_v=x'(t)$ is the velocity.

The rate of dissipation of energy is the same through time,

$$\Delta P(t) = 0,$$

\noindent with $P(t) = -c  y_v(t)  y_x(t).$

The time series data in the DHO dataset includes various parameters characterising the oscillator's motion: time $t$, displacement $y_x$, velocity $y_v$ and acceleration $y_a=x''(t)$. To generate the dataset, we solved the IVP using the Runge-Kutta method of order 5 of Dormand-Prince-Shampine with varying time intervals and sampling frequencies to accommodate the three different experiments. The details of the available data for each experiment are as follows:
\begin{itemize}
	\item \textbf{reconstruction:} This experiment involves training and testing sets with 400 points equally spaced in the time interval $(0,50)$;
	\item \textbf{extrapolation:} The training set consists of 400 points equally spaced in the time interval $(0,50)$, while the testing set includes 400 equally spaced points in the extended time interval $(0,400)$; 
	\item \textbf{completion:} The training set consists of 400 points equally spaced in the time interval $(0,50)$, while the testing set includes 600 equally spaced points in the same time interval $(0,50)$; 
\end{itemize}

By providing data capturing the dynamics of a damped harmonic oscillator under different conditions, the DHO dataset offers a valuable resource for studying the predictive performance of the models trained with and without the proposed self-adaptive penalty method for mechanical systems subject to damping forces.

The DHO dataset with the training and testing sets to conduct every experiment is publicly available on \emph{Kaggle} \cite{coelho_oscillator_2023}.

\end{appendices}

\bibliographystyle{ACM-Reference-Format.bst}
\bibliography{sn-bibliography}

\end{document}